\documentclass{article}

\usepackage{PRIMEarxiv}

\usepackage[utf8]{inputenc} 
\usepackage[T1]{fontenc}    
\usepackage{hyperref}       
\usepackage{url}            
\usepackage{booktabs}       
\usepackage{amsfonts}       
\usepackage{nicefrac}       
\usepackage{microtype}      
\usepackage{lipsum}
\usepackage{caption}
\usepackage{subcaption}
\usepackage{xcolor}
\usepackage[dvipsnames]{xcolor}
\usepackage{arydshln}
\usepackage{amsmath}
\usepackage{makecell}
\usepackage{fancyhdr}       
\usepackage{graphicx}       
\graphicspath{{media/}}     

\title{Case Study: Transformer-based Solution for the Automatic Digitization of Gas Plants
}

\author{
  Bailo I. \quad Buonora F. \quad Ciarfaglia G. \quad Consoli L.T. \quad Evangelista A. \quad Gabusi M. \\ \textbf{\quad Ghiani M. \quad Petracca Ciavarella C. \quad Picariello F. \quad Sarcina F.\quad Tuosto F.\quad Zullo V.}  \\ 
  Eng AI\&Data @Engineering Group \\
\And 
Airoldi L. \quad Bruno G. \quad Gobbo D.D. \quad Pezzenati S. \quad Tona G.A. \\
Snam S.p.A. \\
}

\begin{document}
\maketitle

\begin{abstract}

The energy transition is a key theme of the last decades to determine a future of eco-sustainability, and an area of such importance cannot disregard digitization, innovation and the new technological tools available. This is the context in which the Generative Artificial Intelligence models described in this paper are positioned, developed by Engineering Ingegneria Informatica SpA in order to automate the plant structures acquisition of SNAM energy infrastructure, a leading gas transportation company in Italy and Europe. 
The digitization of a gas plant consists in registering all its relevant information through the interpretation of the related documentation. Currently, this process is carried out manually using the Modulo Grafico Misura (MGM) application and is a laborious task that requires a lot of resources. The aim of this work is therefore to design an effective solution based on Artificial Intelligence techniques to automate the extraction of the information necessary for the digitization of a plant, in order to streamline the daily work of MGM users. The developed solution receives design data, the equipment list and the Piping \& Instrumentation Diagram (P\&ID) of the plant as input, each one in pdf format, and uses OCR, Vision LLM, Object Detection, Relational Reasoning and optimization algorithms to return an output consisting of 2 set of information: a structured overview of the relevant design data and the hierarchical framework of the plant. To achieve convincing results, we extend a state-of-the-art model for Scene Graph Generation introducing a brand new Transformer architecture with the aim of deepening the analysis of the complex relations between the plant's components. The synergistic use of the listed AI-based technologies allowed to overcome many obstacles arising from the high variety of data, due to the lack of standardization. An accuracy of 91\% has been achieved in the extraction of textual information relating to design data. Regarding the plants topology, 93\% of components are correctly identified and the hierarchical structure is extracted with an accuracy around 80\%. 
\end{abstract}


\section{Introduction}

In recent decades, the energy transition has become a central topic in shaping an environmentally sustainable future. Such a critical field cannot overlook the role of digitization, innovation, and emerging technological tools. Within this context, the Generative Artificial Intelligence models discussed in this paper, developed by Eng AI\&Data, have the key role of automating the acquisition of plant structures belonging to the energy infrastructure of SNAM, a leading gas transportation company in Italy and across Europe. \\
With a fleet of about 11,000 Regulating and Metering plants (ReMi) distributed on the Italian gas transportation network, efficient control and management is essential to manage the paperwork related to the creation of new plants or the changes made on existing plants. This activity is focused on extracting relevant information from plants documents and schematics, such as P\&IDs (Piping \& Instrumentation Diagrams) and components lists. \\
The high heterogeneity and complexity of plant types and related documentation have traditionally imposed highly manual and specialized activities by SNAM experts, who receive the documentation from plant owners, analyze it, extract the necessary information (plant structures, lines, sections, components and related characteristics), and enter the extracted details into SNAM internal IT systems. However, this manual registration is a laborious task and requires a lot of resources. Moreover, it could lead to typos or general inaccuracies, not ensuring consistency. \\
The goal of the solution described in this paper is therefore to make such a data extraction process automatic through Generative Artificial Intelligence tools, specifically by processing plant schematics of varying complexity and type and their accompanying documentation. Time and resources to be spent on this task would decrease drastically and automatic systems are generally accurate in repetitive tasks, allowing to reduce the error rate in the registration. \\
An automatic digitization system should therefore receive the plant documents and analyse them to extract the relevant information. The main obstacle stems from the lack of rigorous standardisation in the drafting of key elements of the documentation, which are therefore characterised by a high degree of variability. As a result, the information to be extracted appears within the documents in different orders, formats and styles, making it impossible for simple systems to read. Solutions must therefore be able to analyse the documentation in a flexible manner, simulating the type of reasoning that a human operator would apply in interpreting the documents. \\
This study proposes an end-to-end system for the digitization of gas plants based on the relevant documentation. This system consists of several progressive steps that analyse the main elements and combine the extracted information, each of which uses advanced computer vision and optimization techniques. Compared to existing solutions, which are based on the analysis of a few key elements compared to those available, this study aims to maximise the accuracy of the system by leveraging the entire documentation.

\section{Background}
Digitization of plants documentation could involve several analisys tasks, such as text, tables and diagram recognition and interpretation. Text and table reading is possible through consolidated and evolving techniques like OCR or Vision LLMs. \cite{karmanov2025eclairextractingcontent} \cite{zhou2024enhancing}. \\ 
An important reference element of plant documentation is the P\&ID, a detailed diagram showing installed equipment and their connections, used in all the processes on the plant, from installation to mantainance. The transformation of the P\&ID in a structured representation brings many advantages for the whole plant analisys. A widespread and reasonable method is the encoding of the diagram in a Scene Graph containing information on present components and their relations. \\ 
Existing methods to perform Scene Graph Generation (SGG) from an image are divided into two-stage and one-stage methods: two-stage methods \cite{yang2018graph} first identify entities in the scene to then predict relations between them, while one-stage methods \cite{im2024egtr} \cite{li2022sgtr} \cite{teng2022structured} directly generate subject-predicate-object candidate triplets. The former are less efficient, as they are more onerous, slower and more prone to error propagation, thus the latter represent the state of the art in this field. \\
Within the plant digitization field, several works used combinations of the proviously described techniques to develop smart solutions. \\
S.O. Kang et.al. \cite{kang2019digitization} proposed a tool for the conversion of paper P\&ID into a digital format. The system is based on a pipeline that includes OCR algorithms for text recognition and CNN for symbols detection. A distinctive aspect of this work is its integration with industrial components databases, allowing the mapping of the detected elements with the associated technical specifications. \\
S.Mani et.al. \cite{mani2020automatic} combined computer vision and graph research techniques. They use a CNN for main symbols detection, merge text and symbols information and later leverage a graph research algorithm to identify simbols connections through diagram's lines. \\
B.C. Kim et.al. \cite{kim2022end} proposed an end-to-end framework for P\&ID digitization combining deep learning techniques and graphic analisys to obtain structured representations of the diagrams. The system has been tested on industrial datasets and proved robustness in components detections and it has the capability of building digital formats compatible with plants-management softwares. \\ 
S. Paliwal et.al. \cite{paliwal2021digitize} proposed an automatic digitization system based on Computer Vision and Machine Learning techniques to extract structured information from P\&ID. It uses a combination of CNN for symbols detections and rule-based models to identify connections, allowing diagrams conversion in a standard digital format. \\
S.M. Gajbhiye et.al. \cite{gajbhiye2023advancing} set YOLOv5, the fifth edition of one of the most advanced architectures for object detection, at the center of their work. The system has been trained on annotated P\&ID datasets and optimized to reach high-precision symbols detection. Compared to traditional methods, YOLOv5 stands out for its capability of quickly process images and identify precisely even small symbols, making it very suitable for real-time industrial applications. \\
J.M. Stürmer et.al. \cite{marius2024transforming} introduced the use of Transformers, a class of deep learning models tipically employed in Natural Language Processing, for P\&ID digitization. Specifically, the system leverages the Vision Transformer architecture to analyse diagrams structure and identify symbols and connections. Thanks to the Transformers capability of catching spatianl relations between elements, the model extracts information with high accuracy, improving performance compared to traditional CNN-based models. \\
Despite significant progress in this field, current technologies still have some limitations. Although advanced models such as YOLOv5 and Transformers have greatly improved symbol recognition, the accuracy of identification can be compromised by variations in design, such as the presence of manual annotations or differences in style. Challenges also remain regarding the management of connections between components, the ability to generalise to different industry standards, and the integration of the solution with existing software. Finally, most solutions only consider the P\&ID, which, although it contains a lot of condensed information, may not be sufficiently comprehensive. \\

\section{Methods}

\begin{figure}[h]
    \centering
    \includegraphics[width=0.65\linewidth]{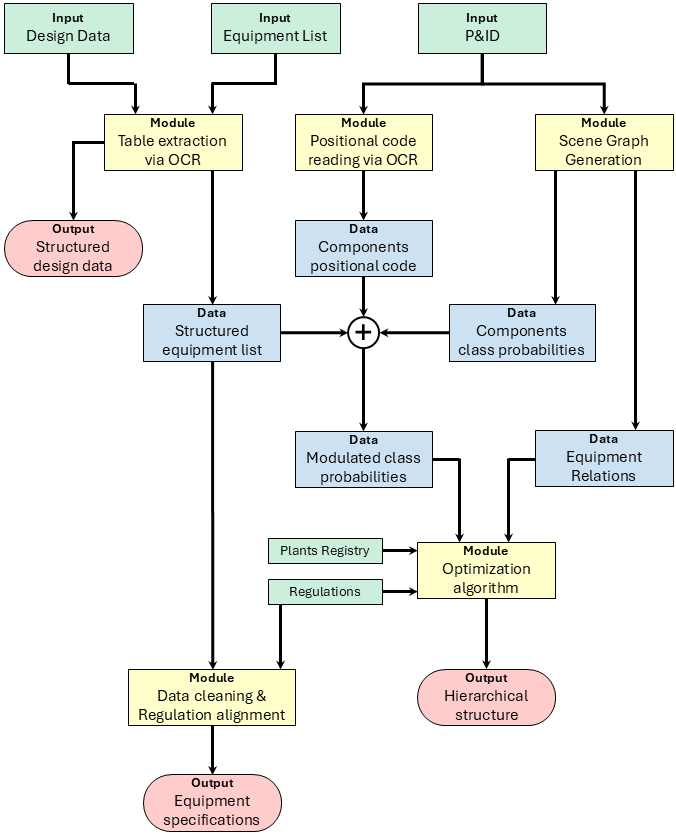}
    \caption{Diagram of the developed solution. All the input data are in pdf format}
    \label{fig:pipeline}
\end{figure}

An overview of the designed pipeline of our solution is provided in Figure \ref{fig:pipeline}. The solution aims to achieve 2 main goals. The first is to extract and organize the relevant design data of the ReMi plant, which is performed through OCR methods. The second and most challenging is to build a hierarchical representation of the plant, which is done using a combination of OCR and LLM methods, SGG models and optimization algorithms in order to maximize the generalization capabilities of the system given the great variability of data. Specifically, the hierarchy tipically used to represent a plant is characterized by the following levels, from the specific to the general:
\begin{itemize}
   \item  Level I: characteristic or \textbf{type}, which serves to define a quantitative or qualitative aspect of a device;
   \item Level II: component or \textbf{sub-type}, which is defined on the basis of a set of mandatory and optional characteristics;
   \item Level III: \textbf{section}, i.e. a coherent grouping of several components that perform a specific function;
   \item Level IV: \textbf{line}, i.e. a set of sections, which may be for regulation or measurement.
\end{itemize}

\subsection{Input Data}
\begin{figure}[h]
    \centering
    \begin{subfigure}[c]{0.247\linewidth}
        \includegraphics[width=\linewidth]{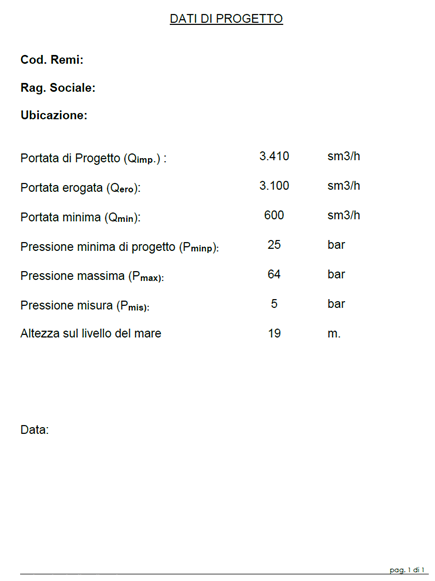}
        \subcaption{}
        \label{fig:dati_progetto}
    \end{subfigure}
    \hfill
    \begin{subfigure}[c]{0.24\linewidth}
        \includegraphics[width=\linewidth]{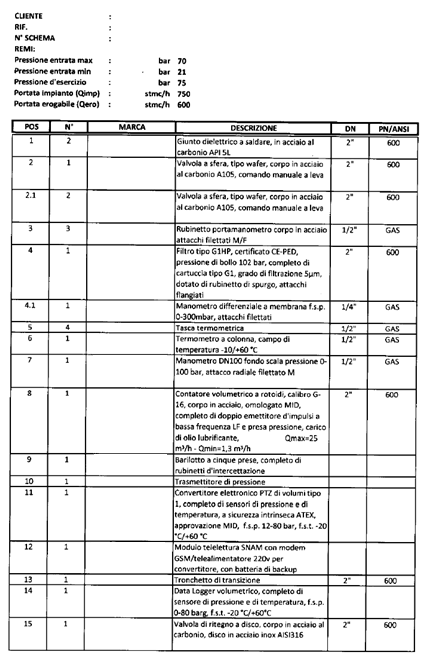}
        \subcaption{}
        \label{fig:apparecchiature}
    \end{subfigure}
    \hfill
    \begin{subfigure}[c]{0.49\linewidth}
        \includegraphics[width=\linewidth]{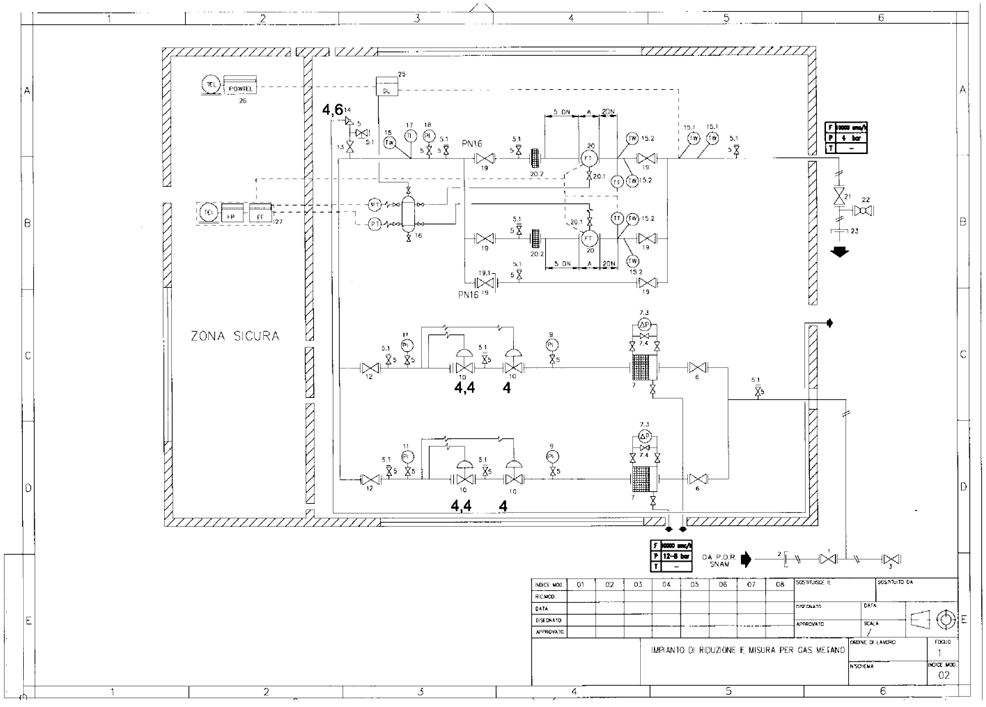}
        \subcaption{}
        \label{fig:PID}
    \end{subfigure}
    \caption{Examples of design data (a), page of equipment list (b) and P\&ID (c)}
    \label{fig:input_data}
\end{figure}

The main input data, examples of which are shown in Figure \ref{fig:input_data}, consist mainly of PDF documents, each corresponding to an aspect of the plant description and including:
\begin{enumerate}
    \item \textbf{Design data} (Figure \ref{fig:dati_progetto}): a page, usually in tabular or similar format, containing general information about the plant, such as its main flow rates and pressures and its height above sea level. Its content must be reported in full in the output;
    \item \textbf{Equipment list} (Figure \ref{fig:apparecchiature}):  one or more pages in tabular format which, with reference to the P\&ID, list the equipment and components present, specifying their identification in the diagram, category, manufacturer, description and main technical specifications;
    \item \textbf{Piping \& Instrumentation Diagram} (Figure \ref{fig:PID}): a page containing a schematic drawing of the plant, showing the components and their connections. Some of the information that can be extracted from the P\&ID and specified in the equipment list constitutes the second part of the output
\end{enumerate}
The plant documentation dataset, extracted and provided by SNAM, consists of approximately 1500 ReMi. \\
Other relevant data that is not provided as input to the pipeline but that has been used for its design and for training some of its models is that concerning the Regulations and the Plant Registry. \\ 
The Regulations consist of a set of Excel files that translate the information in the UNI 9167 standard, the main regulatory reference for the design and management of a plant. These contain, for example, the catalogue of possible equipment, which contains information on the brand, model and characteristics of each possible component that can be used, and the catalogue with the rules for hierarchically dividing the plant into information units. The information contained in the Regulations was used to define sets of rules and constraints that the solution's output must comply with. \\
The Plant Registry is a set of documents containing manually recorded information on all plants installed up to the present. It specifies the characteristics of each plant's components and the hierarchical structure of the plant itself, data that has been used to extract a statistical representation of the composition of the plants.

\subsubsection{Synthetic Data Generation Using Diffusion Models and Algorithmic Techniques}
\begin{figure}[h]
    \centering
    \includegraphics[width=0.8\linewidth]{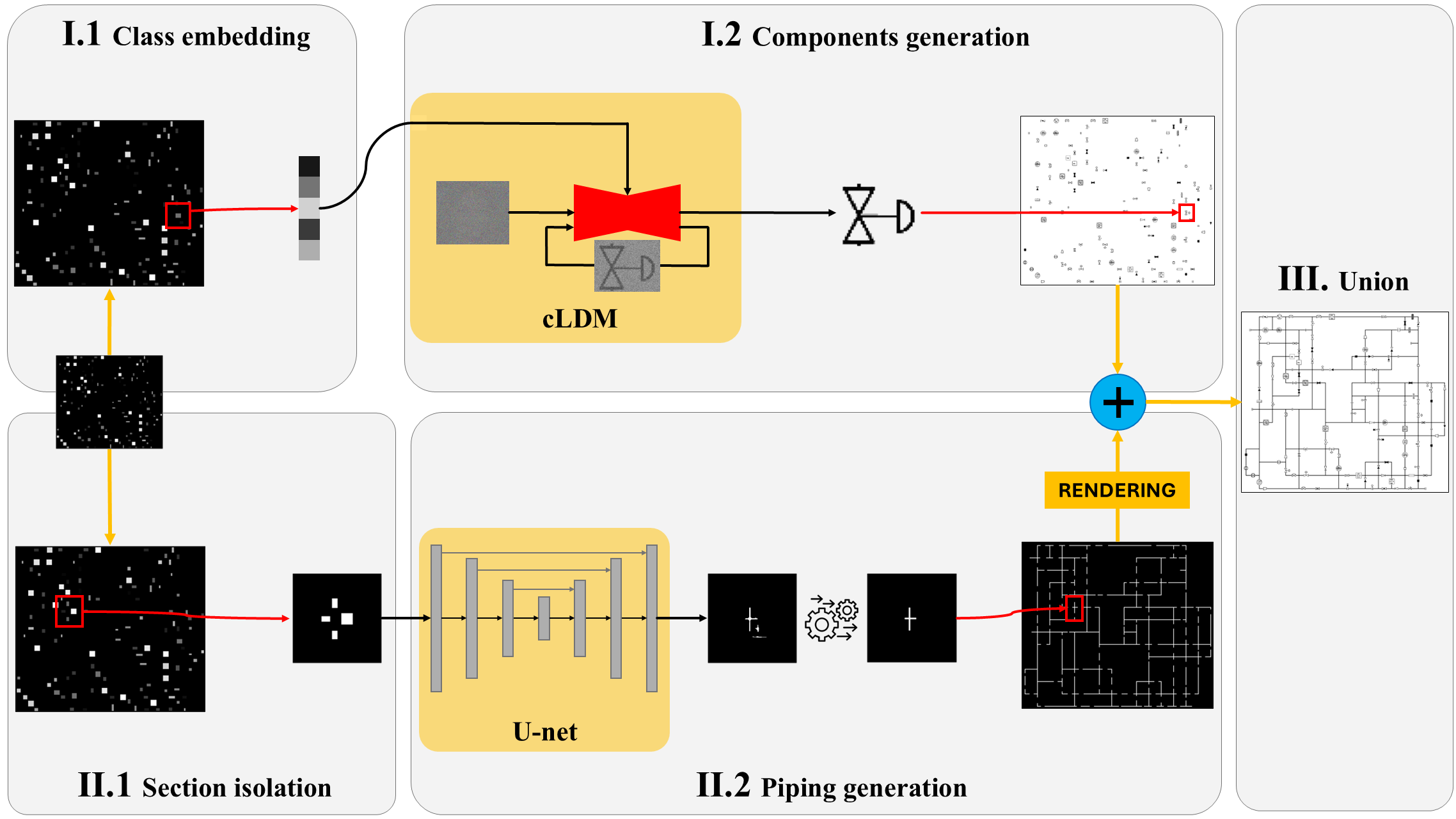}
    \caption{Developed pipeline for synthetic data generation}
    \label{fig:synth_pipeine}
\end{figure}

The availability of P\&ID diagrams represents a significant obstacle to the optimization of the system capabilities, especially that of the SGG model. The amount of original data is insufficient to ensure adequate generalisation capacity, given the high level of diversification of the diagrams. To achieve satisfactory performance, it was therefore necessary to use data augmentation techniques to increase the cardinality of the available data. Synthetic data was added using both algorithmic approaches and based on Generative AI. \\
The first approach consists of the "manual” production of images containing symbols of components present in the original diagrams, positioned within the image after applying random transformations such as distortions, rotations and the addition of noise. The image is then modified to increase its noise, for example by adding text, tables and figures or setting random backgrounds. The images produced using this method were mainly used in the pre-training phase of the SGG model, focused on the object detection task. \\
The second approach consists of using a pipeline that produces a fictitious diagram based on its layout and information regarding the division of symbols into sections. The generative process, schematized in Figure \ref{fig:synth_pipeine}, is divided into two steps. In the first step, using the classes and positions of each component identifiable from the layout, a conditional Latent Diffusion Model is used to generate plausible versions of the symbols. In the second step, a U-net receives the image portions with the components divided into sections and generates the connection map for each group, which is then rendered.

\subsection{Plant's Design Data and Equipment List Extraction}

The first functional block of the entire pipeline consists of extracting design data from the relative documentation page and the specifications of the components present in the plant as listed in the equipment list. \\
The first step in the process is to use an OCR model to extract all tables from the documentation in text format. The obtained output is then processed using an LLM to get ordered and structured tables. The operations performed by the LLM include:
\begin{itemize}
    \item \textbf{Classification} of extracted tables: only those containing the information of interest are selected;
    \item \textbf{Data cleaning and restructuring}: using ad hoc prompts, the filtered tables are purged of irrelevant content and converted into a consistent tabular format;
    \item \textbf{Alignment with Regulations}: a comparison is made between the extracted components and those required by technical regulations in order to identify the mandatory characteristics for each item on the equipment list. These characteristics are then extracted, through further prompts, from the description and, where present, from other columns of the table.
\end{itemize}
The procedure described allows the automatic generation of the plant design data table, which also includes all the components and their respective characteristics required by the Regulations, specifically regarding levels I e II of the hierarchy. This approach overcomes the difficulties associated with the heterogeneity of formats and table structures in technical plant documents.

\subsection{Plant's Hierarchical Structure Reconstruction}
In this phase, the analysis of the plant's P\&ID plays a central role, and it is divided into three phases. In the first phase, an SGG model used to extract an indirect graph from the diagram. This graph identifies the symbols corresponding to the equipment of interest, including their locations and types, and traces the connections between them, allowing for an initial grouping into sections. In the second phase, OCR techniques are used to find the numerical codes associated with each symbol within the diagram. By subsequently identifying the correspondence in the equipment list table, extracted in the task described in section 3.2, it is possible to correct any inaccuracies in the SGG model and finally merge the results. In the last phase, an optimization algorithm uses the merged output to reconstruct the whole hierarchical structure of the plant. \\
Given a set of components \( \mathcal{C} = \{ c_1, c_2, \dots, c_N \} \), it's possible to define sections \( \mathcal{T} = \{ T_1, \dots, T_K \} \) as a partition of \( \mathcal{C} \), i.e. as subsets of \( \mathcal{C} \) so that:
\[
T_k \subseteq \mathcal{C}, \quad T_i \cap T_j = \emptyset \ \forall i \ne j, \quad \bigcup_{k=1}^K T_k = \mathcal{C}.
\]
Similarly, it's possible to define lines \( \mathcal{L} = \{ L_1, \dots, L_M \} \) as a partition of sections \( \mathcal{T} \):
\[
L_j \subseteq \mathcal{T}, \quad L_j \cap L_{j'} = \emptyset \ \forall j \ne j', \quad \bigcup_{j=1}^M L_j = \mathcal{T}.
\]
Each component \( c_i \in \mathcal{C} \), section \( T_k \in \mathcal{T} \) and line \( L_j \in \mathcal{L} \) is then associated a class that defines its type, \( y_i \in \mathcal{Y}_C \), \( y_{T_k} \in \mathcal{Y}_T \), \( y_{L_j} \in \mathcal{Y}_L \) respectively. The hierarchical structure of the plant is then defined by the triplet \((\mathcal{T}, \mathcal{L}, \mathbf{y})\), where \( \mathbf{y} = \{ y_i \}_{i=1}^N \cup \{ y_{T_k} \}_{k=1}^K \cup \{ y_{L_j} \}_{j=1}^M \) is the complete assignment of classes.

\subsubsection{Scene Graph Generation} 
\begin{figure} [h]
    \centering
    \includegraphics[width=0.8\linewidth]{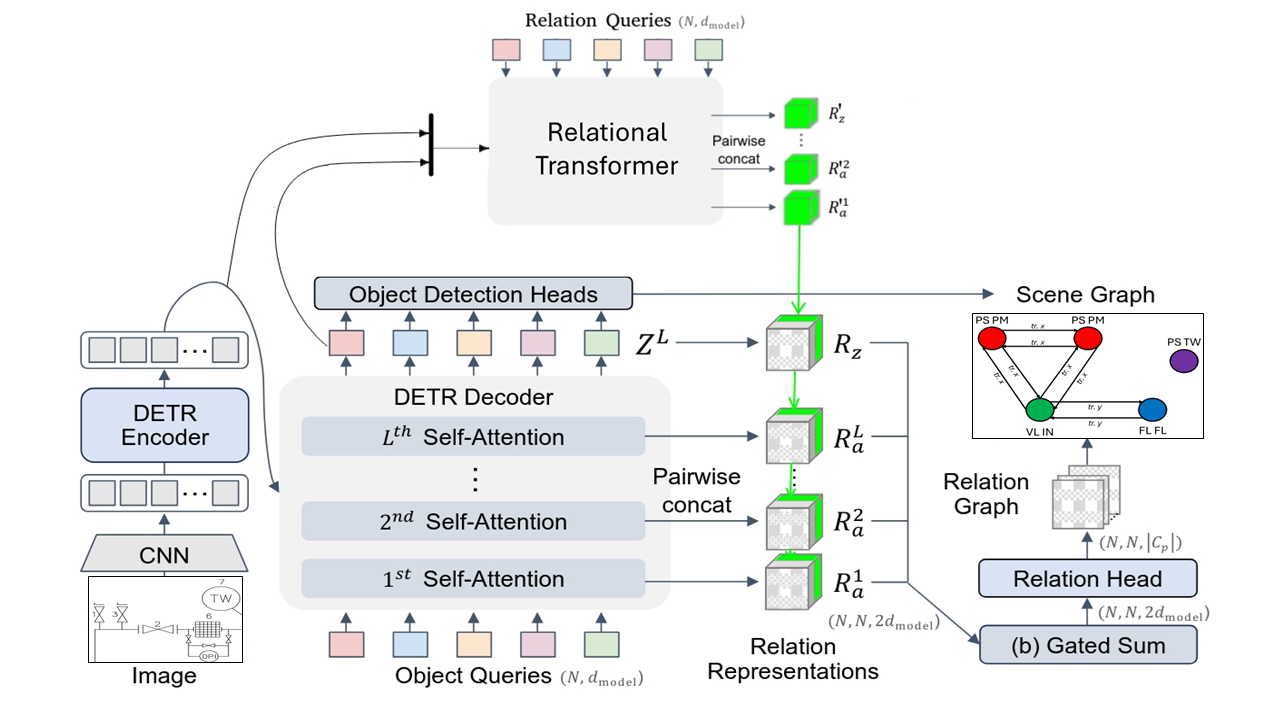}
    \caption{EGRTR architecture}
    \label{fig:egrtr}
\end{figure}
The extraction of a Scene Graph from the plant's P\&ID provides a first representation of the plant based on the diagram topology. \color{black} To address this challenge, the EGRTR (Extracting Graph from Relational Transformer) model was developed, an innovative architecture for SGG that extends the EGTR framework \cite{im2024egtr} by introducing advanced contextual reasoning capabilities. The architecture, which belongs to the class of one-stage methods and is shown in Figure \ref{fig:egrtr}, combines two approaches: EGTR extracts local relationships from the self-attention layers of the DETR decoder, while the Relational Transformer captures global relationships using specialised queries that observe the entire scene through the encoder's features.

\paragraph{Starting Pipeline}
The architecture begins with the standard DETR pipeline:
\begin{enumerate}
    \item Backbone CNN: receives the P\&ID image and extracts a feature map;
    \item DETR encoder: processes the feature maps by integrating contextual information;
    \item DETR decoder: processes $N$ object queries through $L$ layers of self-attention, generating intermediate ($\{Z^l\}_{l=0}^{L-1}$) and final ($Z^L$) representations.
\end{enumerate}

\paragraph{Relational Reasoning}
The extraction of relationships is based on two main flows. \\
In the EGTR flow, which focuses on local analysis, interactions between objects that emerged during the detection process are captured. For each layer $l$ of the DETR decoder, queries and keys are projected and replicated to form specific $R^l_a$ relationship matrices for each pair: 
$$
Q^l_\text{proj} = \text{Proj}_Q(Q^l), \quad K^l_\text{proj} = \text{Proj}_K(K^l)
$$
$$
Q^l_\text{proj} = \text{Proj}_Q(Q^l), \quad K^l_\text{proj} = \text{Proj}_K(K^l)$$ $$R^l_a[i,j,:] = [Q^l_\text{proj}[i,:]; K^l_\text{proj}[j,:]] \in \mathbb{R}^{N \times N \times 2d_\text{model}}
$$
The final representations are extracted from the last layer ($L$) of the decoder:
$$
R_z[i,j,:] = [\text{Proj}_\text{sub}(Z^L[i,:]); \text{Proj}_\text{obj}(Z^L[j,:])] \in \mathbb{R}^{N \times N \times 2d_\text{model}}
$$ 
The Relational Transformer, focused on global context analysis, processes features from the encoder and decoder. Initially, it enriches them through the projection of sinusoidal and learnable positional encoding (PE):
$$
F_\text{enc}^* = \text{Proj}_\text{enc}(\text{PE}_\text{enc}(F_\text{enc})) \quad \quad Z^{L*} = \text{Proj}_\text{dec}(\text{PE}_{dec}(Z^L))
$$ 
$$
F_\text{combined} = [F_\text{enc}^*; Z^{L*}] \in \mathbb{R}^{(S+N) \times d_\text{model}}
$$
Then, a set of $N$ relational queries $\mathcal{Q}_\text{rel} \in \mathbb{R}^{N \times d_\text{model}}$ is processed by a dedicated Transformer Decoder.
$$
\mathcal{Q}'_\text{rel} = \text{RelTransformer}(\mathcal{Q}_\text{rel}, F_\text{combined}) \in \mathbb{R}^{B \times N \times d_\text{model}}
$$
The Relational Transformer then applies:
\begin{itemize}
    \item Self attention to specialize each expert: $\mathcal{Q}_\text{rel} \leftarrow \text{SelfAttn}(\mathcal{Q}_\text{rel}, \mathcal{Q}_\text{rel}, \mathcal{Q}_\text{rel})$;
    \item Cross attention to extract global patterns: $\mathcal{Q}_\text{rel} \leftarrow \text{CrossAttn}(\mathcal{Q}_\text{rel}, F_\text{combined}, F_\text{combined})$.
\end{itemize}
From the total $N$ queries processed $\mathcal{Q}'_{rel}$, the first $L+1$ experts ($\mathcal{E}$) are selected and each is replicated for all pairs of objects
$$
R^{'l}_a[i,j,:] = \mathcal{E}[l,:] \quad \forall i,j \in [1,N], \forall l \in [0,L]
$$

\paragraph{Multi-Expert Adaptive Fusion}
For each layer, the representations extracted from the two flows are concatenated:
$$
\tilde{R}^l = [R^l_a; R'^{l}_a] \in \mathbb{R}^{N \times N \times 3d_\text{model}} \quad \forall l \in [0, L-1]
$$ 
$$
\tilde{R}^{L} = [R_z; R'^{L}_a] \in \mathbb{R}^{N \times N \times 3d_\text{model}}
$$
For each pair of objects $(i,j)$ and each layer/expert $l$, a custom gate determines its importance:
$$
g^l_{i,j} = \sigma(\text{MLP}_\text{gate}(\tilde{R}^l[i,j,:])) \in \mathbb{R}^{1}
$$
All representations are therefore merged through the weights of the relative gates:
$$
\mathcal{R}_\text{fused}[i,j,:] = \sum_{l=0}^{L} g^l_{i,j} \odot \tilde{R}^l[i,j,:] \in \mathbb{R}^{N \times N \times 3d_\text{model}}
$$
Finally, the result is converted into the final graph:
$$
\hat{G}_\text{rel} = \sigma(\text{MLP}_\text{rel}(\mathcal{R}_\text{fused})) \in \mathbb{R}^{N \times N \times |\mathcal{Y}_T|}
$$ 
$$
\hat{G}_\text{conn} = \sigma(\text{MLP}_\text{conn}(\mathcal{R}_\text{fused})) \in \mathbb{R}^{N \times N \times 1}
$$

\paragraph{Benefits}
In the context of P\&ID analysis, EGRTR allows certain complexities to be overcome. Experts are able to cover local ($l\sim1$) and global ($l\sim L$) aspects, capturing both short-range and macroscopic details and relationships. As a result, local experts ensure accuracy in fine details, while the context analysis provided by global experts allows for overcoming non-trivial challenges: for example, it is possible to effectively distinguish graphically similar symbols such as different types of valves and identify long-range connections suggested by pipes.

\paragraph{Output}
EGRTR produces a multi-component output:
\begin{itemize}
    \item Bounding box and class probability for each identified component;
    \item A Relational Scene Graph $G_\text{rel}$, i.e. an $N \times N \times |\mathcal{Y}_T|$ matrix where $G_\text{rel}[i,j,k]$ represents the probability of the $k$-th relationship between components $i$ and $j$ identified;
    \item A Connectivity Graph $G_\text{conn}$, i.e. an $N \times N \times 1$ matrix where $G_\text{conn}[i,j]$ indicates the probability of direct connection between components $i$ and $j$ identified.
\end{itemize}
The output structured in this way encodes the components identified with their types, positions and characteristics (through the nodes and their attributes), the relationships between them with associated probabilities (through the arcs) and the structure of the P\&ID up to level III (through the graph topology), and constitutes a piece of the input for the final optimization algorithms.

\subsubsection{P\&ID and Equipment List Matching via OCR}
This section takes as inputs the plant's P\&ID and the structured equipment list extracted in the previous task. \\
Here, a further OCR analysis is performed on the image of the diagram in order to identify the numerical codes related to each present symbols, extracting their values and coordinates in the image. At this point, each identified code is associated with two entities:
\begin{itemize}
    \item Its correspondence in the previously structured equipment list, therefore a component class specified by a couple type-subtype;
    \item The nearest symbol identified by the SGG model.
\end{itemize}
For each symbol of the scene graph will be built a new list of class probabilities, related to the interpretation of the equipment list. This new set of values is used to modulate the output class probabilities of the SGG model, in order to combine topological and descriptive information. 

\subsubsection{Optimization Algorithm}
In order to harmonize the final output of the solution with the information contained in the Regulations, post-processing is performed on the hierarchical representation of the P\&ID provided by the SGG model. The Regulations define a set of rules and constraints that the hierarchical structures of the plants must satisfy, and it is therefore necessary for the final output of the solution to be compatible with these constraints. In particular, the hierarchical levels subject to post-processing are levels I, II and III, i.e. the definition of component classes (type and sub-type), their grouping into sections and the types of sections. In addition, at this stage, level IV of the plant's hierarchical structure is reconstructed, i.e. the grouping of sections into lines and the definition of line types (measurement or regulation). However, with regard to the permissible compositions of sections in lines, the Regulation defines minimum requirements, and these often do not allow for the unambiguous identification of the plant's lines. The information contained in the plant registry is therefore also used at this stage. The registry specifies the hierarchical structures of the installed plants, from which it is possible to extract a statistical representation of the composition of the plants and define the likelihood of a given hierarchical structure. To produce the final hierarchical structure of the plant, an optimization algorithm has been implemented. It takes as input the merged information extracted from the previous phases of documentation analysis, the set of rules and constraints in the Regulations, and the overall structural likelihood of the plants learned from the registry. 
The merged information extracted from the previous phases of documentation analysis consist of:
\begin{itemize}
    \item Bounding boxes and class probabilities for each identified component. The bounding boxes are provided by the SGG model while the probabilities are weighted averages between those of the SGG model and those of the equipment list matching method;
    \item The Relational Scene Graph $G_\text{rel}$ and the Connectivity Graph $G_\text{conn}$ as they are produced by the SGG model.
\end{itemize}
The goal of the algorithm is to find partitions \(\mathcal{T}, \mathcal{L}\) and classes \(\mathbf{y}\) so that:
\[
\underset{
\mathcal{T}, \mathcal{L}, \mathbf{y}
}{\operatorname{arg\,max}} \quad
\mathcal{S}(\mathcal{T}, \mathcal{L}, \mathbf{y}),
\]
where the objective function to be maximized is:
$$
\mathcal{S}(\mathcal{T}, \mathcal{L}, \mathbf{y}) = \lambda_1\mathbf{E}_{\text{node}}(\mathbf{y}) + 
\lambda_2\mathbf{E}_{\text{edge}}(\mathcal{T}, \mathbf{y}) + 
\lambda_3\mathbf{E}_{\text{struct}}(\mathcal{T}, \mathcal{L}, \mathbf{y}) - 
\lambda_4\mathbf{E}_{\text{norm}}(\mathcal{T}, \mathcal{L}, \mathbf{y}) - 
\lambda_5\mathbf{E}_{\text{reg}}(\mathcal{T}, \mathcal{L}).
$$
The terms of the objective function represent the probabilities provided by the previous phases about components and connections classes, the overall structural plausibility learned from the plant registry, the penalties resulting from violations of the constraints defined in the Regulations, and a regularization. The terms $\mathbf{E}_{\text{node}}(\mathbf{y})$ and $\mathbf{E}_{\text{edge}}(\mathcal{T}, \mathbf{y})$ are defined based on the class probabilities \( \hat{p}_i(y_i) \) associated with each component \( c_i \), the connectivity graph $\hat{G}_{\text{conn}} \in [0,1]^{N \times N}$ representing the probability that two components are connected, and the relational graph \( \hat{G}_{\text{rel}} \in [0,1]^{N \times N \times |\mathcal{Y}_T|} \) representing the probability that a relationship of type \( k \in \mathcal{Y}_T\) exists between two components:
$$
\mathbf{E}_{\text{node}}(\mathbf{y}) = \sum_{i=1}^{N} \log\hat{p}_i(y_i), \quad\quad\quad\quad
\mathbf{E}_{\text{edge}}(\mathcal{T}, \mathbf{y}) = 
\sum_{\substack{k=1 \\ i, j \in T_{k},\, i \ne j}}^{\vert \mathcal{T} \vert} 
\log \left( \hat{G}_{\text{conn}}[i,j]
\cdot
\hat{G}_{\text{rel}}[i,j,y_{T_k}] \right).
$$
The term $\mathbf{E}_{\text{struct}}(\mathcal{T}, \mathcal{L}, \mathbf{y})$ is the logarithm of a plausibility score for the hierarchical structure of the plant defined on the basis of the information contained in the plant registry. It was constructed by training a probabilistic discriminative model that approximates a monotonically increasing likelihood score with respect to the probability that the hierarchical structure corresponds to a real plant \( P(\mathcal{T}, \mathcal{L}, \mathbf{y} \ | \ \text{Real plant}) \). The term $\mathbf{E}_{\text{norm}}(\mathcal{T}, \mathcal{L}, \mathbf{y})$ penalizes hierarchical structures that are not compatible with the constraints of the Regulations, and is defined as:
$$
\mathbf{E}_{\text{norm}}(\mathcal{T}, \mathcal{L}, \mathbf{y}) = 
- \log \left( 
0.5\cdot \frac{\sum_{k=1}^{\vert \mathcal{T} \vert}\phi_T\left( y_{T_k}, \{ y_i \mid c_i \in T_k \} \right)}{\vert \mathcal{T} \vert}  + 
0.5\cdot \frac{\sum_{j=1}^{\vert \mathcal{L} \vert}\phi_L\left( y_{L_j}, \{ y_{T_k} \mid T_k \in L_j \} \right)}{\vert \mathcal{L} \vert}
\right),
$$
where \( \phi_T \) and \( \phi_L \in [0, 1] \) are continuous scoring functions that evaluate the compliance of segments and lines with the minimum requirements of the Regulation. Finally, the regularization term $\mathbf{E}_{\text{reg}}(\mathcal{T}, \mathcal{L})$ penalizes segments that are too large and an excessive number of partitions: 
$$
\mathbf{E}_{\text{reg}}(\mathcal{T}, \mathcal{L}) = 
\alpha_1 \vert \mathcal{T} \vert +
\alpha_2 \vert \mathcal{L} \vert + 
\alpha_3 \sum_{k=1}^{\vert \mathcal{T} \vert} \vert T_k \vert^2.
$$

\section{Results and Discussions}
Table \ref{tab:testi} shows the performance obtained in extracting textual information from the various documents. The values represent, in order, the accuracy of reading the page containing the design data, reading the tables containing the list of equipment, matching the equipment with the respective records in the regulations, and extracting the technical specifications of the components. 
\begin{table}[h]
\centering
\setlength{\arrayrulewidth}{0.4pt} 
\setlength{\tabcolsep}{10pt} 
\renewcommand{\arraystretch}{2} 

\begin{tabular}{c|c|c|c|c}
\textbf{KPI} & \textbf{Design data} & \textbf{Equipment} & \textbf{Registry match} & \textbf{Specifications} \\
\specialrule{1.5pt}{0pt}{0pt} 
\textbf{Accuracy} & 91\% & 80\% & 84\% & 84\%
\end{tabular}

\vspace{5mm}
\caption{Accuracy values obtained in the extraction of textual information from documentation}
\label{tab:testi}
\end{table}

\begin{figure}[h]
    \centering
    \begin{subfigure}[c]{0.3\linewidth}
        \includegraphics[width=\linewidth]{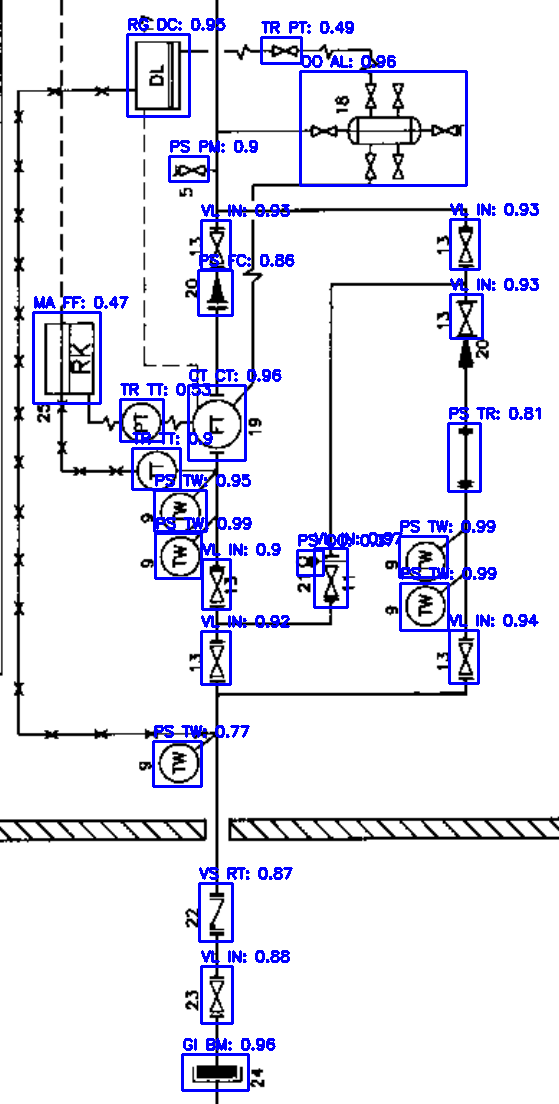}
        \subcaption{}
        \label{fig:ris_DETR}
    \end{subfigure}
    \hfill
    \begin{subfigure}[c]{0.6\linewidth}
        \includegraphics[width=\linewidth]{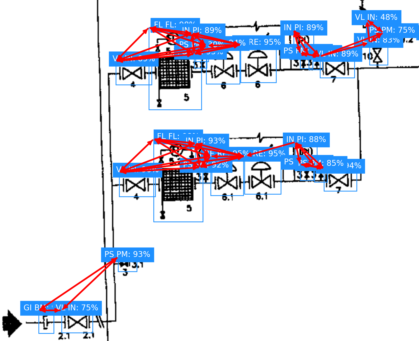}
        \subcaption{}
        \label{fig:ris_EGRTR}
    \end{subfigure}
    \caption{(a) Example of inference by EGRTR for the recognition of components in a test P\&ID. (b) Example of inference by EGRTR for the identification of relationships (in red) between recognised components in a test P\&ID.}
    \label{fig:results}
\end{figure}

Table \ref{tab:egrtr} shows the scores characterizing the final performance of the SGG model. The relatively low values are justified by the following characteristics of the data used:
\begin{itemize}
    \item The bounding boxes of the components, generated manually, are highly variable even for the same symbol. This makes it impossible for the model to accurately predict the coordinates of each bounding box, which is heavily penalized in the mAP calculation as the IoU threshold increases.
    \item Due to the way the dataset was constructed, a single P\&ID can contain a large number of relationships between components, even more than 1000. This is directly reflected in the lowering of traditional scores for measuring the performance of SGG models.
\end{itemize}
However, as far as mAP is concerned, it is more important that the model identifies a component and accurately predicts its class rather than its position. The IoU decision threshold can therefore be relatively low. \\
Furthermore, with regard to recall on relationships, it can be seen that as the number of predicted relationships taken into consideration increases, the score value increases proportionally. This indicates a high accuracy of the model in this field. \\
Visual examples of EGRTR's ability to identify components and relate them to each other based on the section they belong to are shown in Figure \ref{fig:results}. 

\begin{table}[h]
\centering
\setlength{\arrayrulewidth}{0.4pt} 
\setlength{\tabcolsep}{15pt} 
\renewcommand{\arraystretch}{2} 

\begin{tabular}{c|c|c|c|c|c}
\textbf{mAP@[.5]} & \textbf{mAP@[.75]} & \textbf{mAP@[.5:.95]} & \textbf{R@20} & \textbf{R@50} & \textbf{R@100} \\
\specialrule{1.5pt}{0pt}{0pt} 
0.762 & 0.428 & 0.442 & 0.089 & 0.225 & 0.424 
\end{tabular}

\vspace{5mm}
\caption{Main results achieved following EGRTR training on labelled test P\&IDs.}
\label{tab:egrtr}
\end{table}

The above discussion is reflected in the results shown in Table \ref{tab:general_performances}, which relate to the identification of equipment and sections downstream of the optimization algorithm. \\
The percentage value for the components identified is the ratio between those correctly identified and those present in the diagram and available in the ground truth. The correct identification of a component is understood as the correct identification not only of its position but also, and above all, of its type and subtype. \\
The percentage value for the sections is calculated using a weighted average of the individual scores obtained on the sections: this compares how many components in the individual section are correctly associated with the components actually present in the section, weighted by the complexity of the section itself, proportional to the number of components that make it up. The individual scores are then averaged to obtain a single value on the P\&ID.
\begin{table}[h]
\centering
\setlength{\arrayrulewidth}{0.4pt} 
\setlength{\tabcolsep}{15pt} 
\renewcommand{\arraystretch}{2} 

\begin{tabular}{c|c|c|c|c}
\textbf{KPI} & \shortstack{\textbf{Identified} \\ \textbf{Components}} & \shortstack{\textbf{Components correctly} \\ \textbf{assigned to sections}} & \shortstack{\textbf{Identified} \\ \textbf{Regolation sections}} & \shortstack{\textbf{Identified} \\ \textbf{Measure sections}} \\
\specialrule{1.5pt}{0pt}{0pt} 
\textbf{Accuracy} & 95\% & 90\% & 87\% & 85\% 
\end{tabular}

\vspace{5mm}
\caption{Accuracy values 
obtained in the identification of components and sections}
\label{tab:general_performances}
\end{table}

\newpage
\section{Limitations and Conclusions}
This work proposes an end-to-end system for the digitization of gas plants that optimizes results through the comprehensive use of the relevant documentation that is usually available. The analysis pipeline consists of various sequential blocks based on Artificial Intelligence, OCR and constrained optimisation techniques.
Given its performance, the system developed is already applicable as support for MGM users but has room for improvement to make it even more effective. At present, while excellent results have been achieved on simpler systems, performance on complex systems, which are those for which more resources are used, is not excellent. Possible future developments to overcome this limitation would include an increase in the cardinality of the available dataset and the refinement of data augmentation techniques. \\
In conclusion, the system is an effective resource for the purposes of this work, and its components are potentially scalable to other tasks: methods and models can be used in various other fields of study that involve the analysis of technical documentation in addition to the digitization of plants. 
\section{Acknowledgments} 
Part of this work is framed within the Project "AVANT" – Project no. IPCEI-CL\_0000005 - Application protocol no. 108421 of 14/05/2024 - CUP B89J24002920005 - Grant decree no. 1322 of August 8, 2024 - financed by the European Union – NextGenerationEU (IPCEI Funding).


\bibliographystyle{unsrt}  
\bibliography{references}

@article{kang2019digitization,
  title={A digitization and conversion tool for imaged drawings to intelligent piping and instrumentation diagrams (P\&ID)},
  author={Kang, Sung-O and Lee, Eul-Bum and Baek, Hum-Kyung},
  journal={Energies},
  volume={12},
  number={13},
  pages={2593},
  year={2019},
  publisher={MDPI}
}

@inproceedings{mani2020automatic,
  title={Automatic digitization of engineering diagrams using deep learning and graph search},
  author={Mani, Shouvik and Haddad, Michael A and Constantini, Dan and Douhard, Willy and Li, Qiwei and Poirier, Louis},
  booktitle={Proceedings of the IEEE/CVF Conference on Computer Vision and Pattern Recognition Workshops},
  pages={176--177},
  year={2020}
}

@inproceedings{paliwal2021digitize,
  title={Digitize-PID: Automatic digitization of piping and instrumentation diagrams},
  author={Paliwal, Shubham and Jain, Arushi and Sharma, Monika and Vig, Lovekesh},
  booktitle={Trends and Applications in Knowledge Discovery and Data Mining: PAKDD 2021 Workshops, WSPA, MLMEIN, SDPRA, DARAI, and AI4EPT, Delhi, India, May 11, 2021 Proceedings 25},
  pages={168--180},
  year={2021},
  organization={Springer}
}

@article{kim2022end,
  title={End-to-end digitization of image format piping and instrumentation diagrams at an industrially applicable level},
  author={Kim, Byung Chul and Kim, Hyungki and Moon, Yoochan and Lee, Gwang and Mun, Duhwan},
  journal={Journal of Computational Design and Engineering},
  volume={9},
  number={4},
  pages={1298--1326},
  year={2022},
  publisher={Oxford University Press}
}

@inproceedings{gajbhiye2023advancing,
  title={Advancing P\&ID digitization with YOLOv5},
  author={Gajbhiye, Shreya M and Bhamre, SR and Tadepalli, LN Teja and Pillai, MR and Uplaonkar, Deepak},
  booktitle={2023 International Conference on Integrated Intelligence and Communication Systems (ICIICS)},
  pages={1--6},
  year={2023},
  organization={IEEE}
}

@article{marius2024transforming,
  title={Transforming Engineering Diagrams: A Novel Approach for P\&ID Digitization using Transformers},
  author={Marius St{\"u}rmer, Jan and Graumann, Marius and Koch, Tobias},
  journal={arXiv e-prints},
  pages={arXiv--2411},
  year={2024}
}

@inproceedings{yang2018graph,
  title={Graph r-cnn for scene graph generation},
  author={Yang, Jianwei and Lu, Jiasen and Lee, Stefan and Batra, Dhruv and Parikh, Devi},
  booktitle={Proceedings of the European conference on computer vision (ECCV)},
  pages={670--685},
  year={2018}
}

@inproceedings{li2022sgtr,
  title={Sgtr: End-to-end scene graph generation with transformer},
  author={Li, Rongjie and Zhang, Songyang and He, Xuming},
  booktitle={proceedings of the IEEE/CVF conference on computer vision and pattern recognition},
  pages={19486--19496},
  year={2022}
}

@inproceedings{teng2022structured,
  title={Structured sparse r-cnn for direct scene graph generation},
  author={Teng, Yao and Wang, Limin},
  booktitle={Proceedings of the IEEE/CVF Conference on Computer Vision and Pattern Recognition},
  pages={19437--19446},
  year={2022}
}

@inproceedings{im2024egtr,
  title={Egtr: Extracting graph from transformer for scene graph generation},
  author={Im, Jinbae and Nam, JeongYeon and Park, Nokyung and Lee, Hyungmin and Park, Seunghyun},
  booktitle={Proceedings of the IEEE/CVF Conference on Computer Vision and Pattern Recognition},
  pages={24229--24238},
  year={2024}
}

@misc{karmanov2025eclairextractingcontent,
      title={\'Eclair -- Extracting Content and Layout with Integrated Reading Order for Documents}, 
      author={Ilia Karmanov and Amala Sanjay Deshmukh and Lukas Voegtle and Philipp Fischer and Kateryna Chumachenko and Timo Roman and Jarno Seppänen and Jupinder Parmar and Joseph Jennings and Andrew Tao and Karan Sapra},
      year={2025},
      eprint={2502.04223},
      archivePrefix={arXiv},
      primaryClass={cs.CV},
      url={https://arxiv.org/abs/2502.04223}, 
}

@article{zhou2024enhancing,
  title={Enhancing Table Recognition with Vision LLMs: A Benchmark and Neighbor-Guided Toolchain Reasoner},
  author={Zhou, Yitong and Cheng, Mingyue and Mao, Qingyang and Liu, Qi and Xu, Feiyang and Li, Xin and Chen, Enhong},
  journal={arXiv preprint arXiv:2412.20662},
  year={2024}
}

\end{document}